\documentclass[preprint,12pt]{elsarticle}

\usepackage{amssymb}
\usepackage{graphicx}%
\usepackage{multirow}%
\usepackage{amsmath,amssymb,amsfonts}%
\usepackage{amsthm}%
\usepackage{mathrsfs}%
\usepackage[title]{appendix}%
\usepackage{xcolor}%
\usepackage{textcomp}%
\usepackage{manyfoot}%
\usepackage{algorithm}%
\usepackage{algorithmicx}%
\usepackage{algpseudocode}%
\usepackage{listings}%
\usepackage{textcomp}
\usepackage{stfloats}
\usepackage{url}
\usepackage{verbatim}
\usepackage{booktabs,multirow}
\usepackage{makecell}
\usepackage{bigstrut}
\usepackage{array}
\usepackage{ragged2e}
\usepackage{hyperref}

\journal{Neurocomputing}

\begin{document}

\begin{frontmatter}


\title{MaxMin-L2-SVC-NCH: A Novel Approach for Support Vector Classifier Training and Parameter Selection}

\author[1,2]{Linkai Luo \corref{cor1}}
\ead{luolk@xmu.edu.cn}
\author[1,2]{Qiaoling Yang}
\author[1,2]{Hong Peng}
\author[1,2]{Yiding Wang}
\author[1,2]{Ziyang Chen}

\address[1]{Department of Automation, Xiamen University, Xiamen, 361102, China}
\address[2]{Xiamen Key Laboratory of Big Data Intelligent Analysis and Decision-Making, Xiamen, 361102, China}

\cortext[cor1]{Corresponding author}

\begin{abstract}
The selection of Gaussian kernel parameters plays an important role in the applications of support vector classification (SVC). A commonly used method is the k-fold cross validation with grid search (CV), which is extremely time-consuming because it needs to train a large number of SVC models. In this paper, a new approach is proposed to train SVC and optimize the selection of Gaussian kernel parameters. We first formulate the training and the parameter selection of SVC as a minimax optimization problem named as MaxMin-L2-SVC-NCH, in which the minimization problem is an optimization problem of finding the closest points between two normal convex hulls (L2-SVC-NCH) while the maximization problem is an optimization problem of finding the optimal Gaussian kernel parameters. A lower time complexity can be expected in MaxMin-L2-SVC-NCH because CV is not needed. We then propose a projected gradient algorithm (PGA) for the training of L2-SVC-NCH. It is revealed that the famous sequential minimal optimization (SMO) algorithm is a special case of the PGA. Thus, the PGA can provide more flexibility than the SMO. Furthermore, the solution of the maximization problem is done by a gradient ascent algorithm with dynamic learning rate. The comparative experiments between MaxMin-L2-SVC-NCH and the previous best approaches on public datasets show that MaxMin-L2-SVC-NCH greatly reduces the number of models to be trained while maintaining competitive test accuracy. These findings indicate that MaxMin-L2-SVC-NCH is a better choice for SVC tasks.

\end{abstract}


\end{frontmatter}

\section{Introduction}
Support vector classifier (SVC) \cite{1995Support} is one of the most successful machine learning methods. The SVC with Gaussian radian basis function (GRBF) kernel performs well on many classification tasks\cite{vidic2018support,sethy2020deep,sun2017dynamic,PAN201790,hoang2018image,fan2019research,manavalan2017svmqa}. Although SVC is a traditional method, there are still many research on improving SVCs recently \cite{ruan2021convex,avolio2020semiproximal,tang2021valley,li2021dc,zhang2019optimal}. A major disadvantage of SVC is high time cost in the selection of model’s parameters. The model’s parameters include the penalty parameter $C$ for empirical risk and Gaussian kernel parameter $\gamma$. They need to be tuned for specific tasks so that a good performance can be obtained. The commonly used method to tune $C$ and $\gamma$ is the k-fold cross-validation based on grid search (CV) that is extremely time-consuming. In CV, the candidate values of $C$ and $\gamma$ are obtained with grid division on some value intervals and the dataset is randomly divided into $k$ equal-sized subsets. The optimal $(C,\gamma)$ is the one with the highest average cross-validation accuracy. A large number of SVC models needs to be trained to compute the average cross-validation accuracies at each grid-point $(C,\gamma)$. Thus, the CV is extremely time-consuming.

Many methods have been proposed to reduce the time cost of SVC. Wang et al. (2003) \cite{wang2003determination} applied Fisher discriminant function calculated from GRBF to choose $\gamma$ and verified the effectiveness of the selected $\gamma$ on an a synthetic dataset. However, they don’t provide the comparative experiments on real datasets. Tsang et al. (2005) \cite{tsang2005core} proposed core vector machine (CVM) that is faster than the traditional SVC models with close accuracies. However, the CVM doesn’t involve tuning the optimal $(C,\gamma)$, i.e. the method of tuning $(C,\gamma)$ is still CV and the time cost reduction is due to a lower training cost for the SVC model with given $(C,\gamma)$. Sun et al. (2010) \cite{sun2010analysis} proposed a novel method to tune $\gamma$ by maximizing the distance between two classes (DBTC) in the feature space. They showed that the DBTC implicitly takes the between-class separation into account with a normalized kernel function and the DBTC-based methods outperforms the Span bound, the Radius/Margin bound, kernel Fisher discriminant models and the radial basis function network on most of the benchmark datasets. However, the comparison between the DBTC and the commonly used CV is not provided. Menezes et al. (2019) \cite{menezes2019width} applied a density estimation-based approach to tune $\gamma$ by maximizing a dissimilarity function and obtained a close classification performance as the traditional SVC models do. However, the dissimilarity function is very complicated and the maximizing problem is not easy to solve. In addition, the tuning of $\gamma$ is independent of training SVC, i.e. it belongs to the filter approach rather than the more efficient wrapper approach. Akram-Ali-Hammouri et al. (2022) \cite{akram2022fast} proposed the fast support vector classification (FSVC) that obtains a big improvement in training time, especially for large-scale problems. However, the Kappa on small datasets is inferior to that of the traditional SVC, which indicates that the principle of the FSVC still has shortcomings. Evolutionary computation is often used to choose the hyper-parameters of SVC. Friedrichs and Igel (2005) \cite{FRIEDRICHS2005107} applied the covariance matrix adaptation evolution strategy to choose the hyper-parameters of SVC, while Tharwat et al. (2017) \cite{tharwat2017ba} utilized Bat algorithm to optimize the parameters of SVC. However, the values of the fitness functions are calculated by training SVC models, i.e. a large number of SVC models still need to be trained and the time cost is still huge.

Although there exist many methods as described above in reducing the time cost of SVC, they have shortcomings on classification performance. As far as we know that CV is still the most commonly used method in the selection of hyper-parameters. In this paper, we propose a new method to train SVC including the selection of hyper-parameters so that the time cost can be significantly reduced and the classification performance is not inferior to the traditional method.

Peng et al. (2011) \cite{peng2011soft} proposed a soft-margin support vector model based on normal convex hulls (L2-SVC-NCH) for binary classification. The traditional soft-margin SVC \cite{bennett2000duality}, SVC-RCH, is derived by finding the closest points between two reduced convex hulls (RCHs). Compared to normal convex hull (NCH), RCH is hard to understand. However, the soft-margin SVC in L2-SVC-NCH is modeled by finding the closest points between the positive NCH and the negative NCH where the incomprehensible RCH is removed. Therefore, L2-SVC-NCH has a good intuitive geometric interpretation. The distance of the closest points is just the distance between the positive NCH and the negative NCH. The larger the distance of the closest points, the better generalization performance of the SVC. It indicates that tuning the hyper-parameters of SVC can be considered as maximizing the distance of the closest points of the two NCHs. Therefore, we model the training of SVC and the selection of hyper-parameters as a minimax optimization problem (MaxMin-L2-SVC-NCH), in which the minimization problem is to find the closest points between the two NCHs, i.e. L2-SVC-NCH, while the maximization problem is to find the hyper-parameters that maximize the distance of the closest points. A low time cost can be expected in MaxMin-L2-SVC-NCH because CV is not needed. To solve MaxMin-L2-SVC-NCH quickly, a projected gradient algorithm (PGA) is proposed for training L2-SVC-NCH while a gradient ascent algorithm with dynamic learning rate (GA-DLR) is used in the solution of the maximization problem.  

The major contributions of this paper are as follows.
\begin{enumerate}
\item{The training and the selection of Gaussian kernel parameters of SVC is modeled as a minimax problem, MaxMin-L2-SVC-NCH, in which CV is not needed.}
\item{The PGA is proposed for training the minimization problem L2-SVC-NCH. It is revealed that the famous sequential minimal optimization (SMO) algorithm is a special case of the PGA and the PGA can provide more flexibility.}
\item{The GA-DLR algorithm is proposed for the solution of the maximization problem so that the optimal Gaussian kernel parameters can be obtained by gradient-based methods.}
\item{MaxMin-L2-SVC-NCH greatly reduces the time complexity while maintaining a competitive classification performance compared to the previous best approaches, which indicates that it is a better choice for SVC tasks.}
\end{enumerate}

The remainder of this paper is organized as follows. Section \uppercase\expandafter{\romannumeral2} provides a concise introduction to L2-SVC-NCH. In Section \uppercase\expandafter{\romannumeral3}, we firstly propose MaxMin-L2-SVC-NCH as an innovative approach for training SVC and selecting kernel parameters. We then present the PGA and the GA-DLR algorithm for the solutions of the minimization problem and the maximization problem respectively. A gradient-based algorithm is finally provided for the solution of MaxMin-L2-SVC-NCH by connecting the PGA and the GA-DLR in series. To illustrate that the SMO algorithm is a special case of the PGA, Karush-Kuhn-Tucker (KKT) conditions and the SMO algorithm of L2-SVC-NCH, as well as the comparison between the SMO and the PGA are also provided in Section \uppercase\expandafter{\romannumeral3}. Experimental results and discussions are presented in Section \uppercase\expandafter{\romannumeral4}. Finally, in Section \uppercase\expandafter{\romannumeral5}, we conclude the paper and provide some potential directions for future research.

\section{Related work}
Given a training set $ T=\{ (\boldsymbol{x}_i,\boldsymbol{y}_i) | \boldsymbol{x}_i \in\mathbf{R}^n,  \boldsymbol{y}_i\in\{+1, -1\}, i = 1,...,l\}$, L2-SVC-NCH \cite{peng2011soft} is

\begin{equation}
\label{formula_ex1}
\underset{\boldsymbol{\alpha}}{\rm{min}~}{\frac{1}{2}\boldsymbol{\alpha}^{T}\left( {\boldsymbol{G} + \frac{\boldsymbol{I}}{C}} \right)\boldsymbol{\alpha}}
\end{equation}
\begin{center}
$s.t.~~{\sum\limits_{i \in {ID}^{+}}\boldsymbol{\alpha}_{i}} = {\sum\limits_{i \in {ID}^{-}}\boldsymbol{\alpha}_{i}} = 1,~\boldsymbol{\alpha}_{i} \geq 0,~i = 1,2,...,l$
\end{center}
where $\boldsymbol{G}=[ \boldsymbol{y}_i \boldsymbol{y}_j k(\boldsymbol{x}_i,\boldsymbol{x}_j)]_{l\times l}, k(\cdot,\cdot)$ is a given kernel function, $\boldsymbol{I}$ is is the unit matrix, $C$ is the penalty parameter to empirical risk,  $ ID^+$ and  $ ID^-$  are the index sets of positive samples and negative samples respectively, i.e., 
\begin{align}
{ID}^{+} &= \left\{ i \middle| {\boldsymbol{y}_{i} = 1,~i = 1,2,...,l} \right\},\\
~{ID}^{-} &= \left\{ i \middle| {\boldsymbol{y}_{i} = - 1,~i = 1,2,...,l} \right\}.
\end{align}

The constraints in \eqref{formula_ex1} represents two normal convex hulls. Thus, L2-SVC-NCH can be viewed as an optimal problem of finding the closest points between two normal convex hulls regardless of whether the training set is linearly or non-linearly separable. However, the optimal problem in SVC-RCH is finding the closest points between two reduced convex hulls when the training set is non-linearly separable.    

Suppose  $\boldsymbol{\alpha}^*$ is an optimal solution of \eqref{formula_ex1}, then the decision function for the binary classification problem is
\begin{equation}
\label{formula_ex4}
y(x) = \mathrm{sign}( {{\sum\limits_{i \in S}{\boldsymbol{y}_{i}\boldsymbol{\alpha}_{i}^{*}k(\boldsymbol{x},\boldsymbol{x}_{i}}}) - {( p^{*} + q^{*})/2}})
\end{equation}
where
\begin{equation}
\label{formula_ex5}
p^{*} = {\sum\limits_{i \in S}{\boldsymbol{y}_{i}\boldsymbol{\alpha}_{i}^{*}k\left( {\boldsymbol{x}_{j},\boldsymbol{x}_{i}} \right)}} + {{\boldsymbol{y}_{j}\boldsymbol{\alpha}_{j}^{*}}/C}~\left( {{\exists~\boldsymbol{\alpha}}_{j}^{*} > 0,\boldsymbol{y}_{j} = 1} \right),
\end{equation}
\begin{equation}
\label{formula_ex6}
q^{*} = {\sum\limits_{i \in S}{\boldsymbol{y}_{i}\boldsymbol{\alpha}_{i}^{*}k\left( {\boldsymbol{x}_{j},\boldsymbol{x}_{i}} \right)}} + {{\boldsymbol{y}_{j}\boldsymbol{\alpha}_{j}^{*}}/C}~\left( {\exists~\boldsymbol{\alpha}}_{j}^{*} > 0,\boldsymbol{y}_{j} = - 1 \right).
\end{equation}
$S$ is the collection of support vectors, i.e.
$S = \left\{ i \middle| {\boldsymbol{\alpha}_{i}^{*} > 0,~i = 1,2,...,l} \right\}$\cite{peng2011soft}.

$\boldsymbol{G}$ is often called the gram matrix of kernel function $k(\cdot,\cdot)$. In fact, $(\boldsymbol{G}+\frac{\boldsymbol{I}}{C})$ can be viewed as the gram matrix of $\overset{-}{k}(\cdot,\cdot)$, where
\begin{equation}
\label{formula_ex61}
{\overline{k}}\left( {\boldsymbol{x}_{i},\boldsymbol{x}_{j}} \right) = k\left( {\boldsymbol{x}_{i},\boldsymbol{x}_{j}} \right) + \frac{1}{C}\delta_{ij}
\end{equation}
and $\delta_{ij}$ is Kronecker delta kernel function, i.e.,
\begin{center}
$~\delta_{ij} = \left\{ \begin{matrix}
{1,~~~if~i = j}, \\
{0,~~~if~i \neq j}. \\
\end{matrix} \right.$
\end{center}

L2-SVC-NCH has many advantages compared with traditional SVCs. Table \ref{tab:tabel1} summarizes the advantages of L2-SVC-NCH \cite{peng2011soft}.

\begin{table}[htbp]
  \centering
  \caption{The advantages of L2-SVC-NCH}
    \begin{tabular}{|c|c|}
    \hline
    No &  Description\bigstrut\\
    \hline
    1  & \multicolumn{1}{|p{27em}|}{L2-SVC-NCH can be viewed as finding the closest points between two normal convex hulls and has an intuitive geometric interpretation, while the traditional SVC-RCH is regarded as finding the closest point between two reduced convex hulls and the reduced convex hulls are hard to understand.} \bigstrut\\
    \hline
    2     & \multicolumn{1}{|p{27em}|}{The penalty parameter $C$ can be viewed as a trade-off between the ordinary kernel function and Kronecker delta kernel function in L2-SVC-NCH so that the empirical risk and the confidence risk can be uniformly processed by kernel functions, while the empirical risk cannot be viewed as a kernel function in the traditional SVC-RCH.} \bigstrut\\
    \hline
    3     & \multicolumn{1}{|p{27em}|}{L2-SVC-NCH is a strictly convex quadratic programming problem and has a unique optimal solution, while the traditional SVC-RCH is a convex quadratic programming problem, and it may have a lot of optimal solutions. }\bigstrut\\
    \hline
    \end{tabular}%
  \label{tab:tabel1}%
\end{table}%

\section{Proposed method}
In this section, we firstly formulate the training and parameter selection of L2-SVC-NCH as the minimax problem MaxMin-L2-SVC-NCH. Subsequently, we present the PGA for the solution of the minimization problem after the KKT conditions to L2-SVC-NCH are derived. Additionally, it is revealed that the famous SMO algorithm is a special case of the PGA by a rigorous comparison between them. Moreover, the GA-DLR algorithm is proposed for the solution of the maximization problem. Through connecting the PGA and the GA-DLR in series, a gradient-based algorithm is finally provided for the solution of MaxMin-L2-SVC-NCH. 
\subsection{The minimax problem}

It is necessary to choose the appropriate kernel function and the trade-off parameter \textit{C} between the ordinary kernel function and the Kronecker delta kernel function in the application of L2-SVC-NCH. Considering Gaussian kernel is the most commonly used kernel function, we choose Gaussian kernel as the kernel function. In addition, the parameter \textit{C} is set to a constant since it plays a small role due to the introduction of Gaussian kernel. The reason why $C$ plays a small role when Gaussian kernel is introduced is as follows. Firstly, we can always select a Gaussian kernel so that two-classes problem is linearly separable in the mapped space, while $C$ is not needed for linearly separable problem since empirical risk can be zero by selecting a suitable hyperplane. Further, the distances in the mapped space for $\overline{k}\left( x_{i},x_{j} \right)$ (the kernel function with $C$) and $k\left( x_{i},x_{j} \right)$ (the kernel function without $C$) are $\overline{d}^{2}\left( x_{i},x_{j} \right) = 2 + 2/C - 2k\left( x_{i},x_{j} \right)$ and $d^{2}\left( x_{i},x_{j} \right) = 2 - 2k\left( x_{i},x_{j} \right)$ respectively. Thus, there is no essential difference between $\overline{d}^{2}\left( x_{i},x_{j} \right)$ and $d^{2}\left( x_{i},x_{j} \right)$ because $\overline{d}^{2}\left( x_{i},x_{j} \right)$ can be viewed as a translation of with 2/C when $i \neq j$ while they are the same when $i=j$. In fact,
\begin{equation}
\begin{aligned}
d^2(x_i, x_j) &= (\Phi(x_i) - \Phi(x_j) )^T (\Phi(x_i) - \Phi(x_j)) \\
&= (\Phi(x_i))^T \Phi(x_i) + (\Phi(x_j))^T\Phi(x_j) - 2(\Phi(x_i))^T \Phi(x_j) \\
&= k(x_i, x_i) + k(x_j, x_j) - 2k(x_i, x_j) \\
&= 2 - 2k(x_i, x_j)
\end{aligned}
\end{equation}
where $\Phi$ is the corresponding nonlinear mapping for $k(x_i, x_j)$. In the same way, we have
\begin{equation}
\begin{aligned}
\overline{d}^2 (x_i, x_j) &= \overline{k}(x_i, x_i) + \overline{k}(x_j, x_j) - 2\overline{k}(x_i, x_j) \\
&= 2 + 2/C - 2\overline{k}(x_i, x_j) \\
&= \left\{{\begin{matrix}
        0, & \text{if } i=j \\
    2 + 2/C - 2k(x_i, x_j), & \text{if } i \neq j
\end{matrix}~.}\right.
\end{aligned}
\end{equation}
Menezes et al.\cite{menezes2019width} also verifies that $C$ is unimportant by experiments when Gaussian kernel is introduced.

L2-SVC-NCH is interpreted as finding the closest points between two normal convex hulls. If the kernel function and \textit{C} is given, L2-SVC-NCH is a minimization problem. However, the choice of kernel function is a maximization problem since its goal is to find the closest points with the largest distance. In the corresponding space for the kernel function $\overline{k}\left( x_{i},x_{j} \right)$, the training data must be linearly separable because the distance of the two NCHs is greater than zero. In linearly separable case, a large distance means a better generalization performance. Therefore, training L2-SVC-NCH with the choice of Gaussian kernel can be modeled as a minimax problem, i.e.
\begin{equation}
\label{formula_ex7}
{\underset{\gamma}{\rm{max}~}\underset{\boldsymbol{\alpha}}{\rm{min}~}}{\frac{1}{2}\boldsymbol{\alpha}^{T}\left\lbrack {\boldsymbol{y}_{i}\boldsymbol{y}_{j}e^{- \gamma{\|{\boldsymbol{x}_{i} - \boldsymbol{x}_{j}}\|}^{2}} + \frac{1}{C}\delta_{ij}} \right\rbrack_{l \times l}\boldsymbol{\alpha}}
\end{equation}
\begin{center}
$s.t.~~{\sum\limits_{i \in {ID}^{+}}\boldsymbol{\alpha}_{i}} = {\sum\limits_{i \in {ID}^{-}}\boldsymbol{\alpha}_{i}} = 1,~0 \leq \boldsymbol{\alpha}_{i} \leq 1,i = 1,2,\cdots,l$
\end{center}
where $\gamma$ is the parameter of Gaussian kernel. For convenience, we abbreviate the model \eqref{formula_ex7} as MaxMin-L2-SVC-NCH.

In the traditional SVC, the k-fold cross-validation based on grid search is commonly used to the choice of model parameters, which is extremely time-consuming. Suppose the number of grid points is $m$, it needs training $m\times k$ SVC models to obtain suitable model parameters. However, the choice of model parameters in \eqref{formula_ex7} is modeled a maximization problem, which can be solved by gradient-based algorithms.

Let
\begin{equation}
\label{formula_ex8}
f\left( {\boldsymbol{\alpha},\gamma} \right) = \frac{1}{2}\boldsymbol{\alpha}^{T}\left\lbrack {\boldsymbol{y}_{i}\boldsymbol{y}_{j}e^{- \gamma{\|{\boldsymbol{x}_{i} - \boldsymbol{x}_{j}}\|}^{2}} + \frac{1}{C}\delta_{ij}} \right\rbrack_{l \times l}\boldsymbol{\alpha}.
\end{equation}
A feasible point $\left( {\boldsymbol{\alpha}^{*},\gamma^{*}} \right)$ is called as a saddle point if satisfies
\begin{equation}
\label{formula_ex9}
f\left( {\boldsymbol{\alpha}^{*},\gamma} \right) \leq f\left( {\boldsymbol{\alpha}^{*},\gamma^{*}} \right) \leq f\left( \boldsymbol{\alpha},\gamma^{*} \right)
\end{equation}
i.e. $\boldsymbol{\alpha}^*$ is the minimum point of $f\left( {\boldsymbol{\alpha},\gamma^{*}} \right)$ while  $\gamma^*$ is the maximum point of $f\left( {\boldsymbol{\alpha}^{*},\gamma} \right)$. From the definition of saddle point, we see that saddle point is a local optimum of \eqref{formula_ex7}. An alternate optimization between $\boldsymbol{\alpha}$ and $\gamma$ based on gradient is provided in Section 3.4 to find the local optimal solution of \eqref{formula_ex7}. 

\subsection{The solution of the minimization problem}
\subsubsection{The KKT conditions}

L2-SVC-NCH is a strictly convex quadratic programming problem. Thus, KKT condition is a necessary and sufficient condition for the optimal solution. The following Theorem 1 provides a KKT condition of the optimal solution for L2-SVC-NCH.

\textbf{Theorem 1} Let $f(\boldsymbol{\alpha})$  is the objective function of L2-SVC-NCH, a KKT condition of the optimal solution $\boldsymbol{\alpha}$ for L2-SVC-NCH is that there are \emph{p} and \emph{q} so that for all {\textit{i}}
\begin{equation}
\label{formula_ex10}
\begin{array}{l}
{\text{if}~0 < \boldsymbol{\alpha}_{i}^{+} < 1,\text{then}~{- \nabla}_{i}f(\boldsymbol{\alpha}) = p,} \\
{\text{if}~\boldsymbol{\alpha}_{i}^{+} = 0,\text{then}~- \nabla_{i}f(\boldsymbol{\alpha}) \leq  p,} \\
{\text{if}~\boldsymbol{\alpha}_{i}^{+} = 1,\text{then}~{- \nabla}_{i}f(\boldsymbol{\alpha}) \geq p,} \\
{\text{if}~0 < \boldsymbol{\alpha}_{i}^{-} < 1,\text{then}~{- \nabla}_{i}f\left( \mathbf{\boldsymbol{\alpha}} \right) = q,} \\
{\text{if}~\boldsymbol{\alpha}_{i}^{-} = 0,\text{then}~{-\nabla}_{i}f(\boldsymbol{\alpha}) \leq q,} \\
{\text{if}~\boldsymbol{\alpha}_{i}^{-} = 1,\text{then}~{-\nabla}_{i}f(\boldsymbol{\alpha}) \geq q} \\
\end{array}
\end{equation}
where $\boldsymbol{\alpha}_i^+$ or $\boldsymbol{\alpha}_i^-$  indicates that the $i$-th sample belongs to positive or negative class, ${\nabla}_{i}f(\boldsymbol{\alpha})$ is the $i$-th component of the gradient ${\nabla}f(\boldsymbol{\alpha})$, and
\begin{equation}
\label{formula_ex11}
\nabla f(\boldsymbol{\alpha}) = (\boldsymbol{G} + \frac{\boldsymbol{I}}{C})\boldsymbol{\alpha}.
\end{equation}

\textbf{Proof}  The Lagrange function of L2-SVC-NCH is
\begin{equation}
\text{$L(\boldsymbol{\alpha},p,q,\boldsymbol{\lambda},\boldsymbol{\beta}) = f(\boldsymbol{\alpha}) + p(( \boldsymbol{e}^{+})^{T}\boldsymbol{\alpha} - 1) +$}
\text{$q((\boldsymbol{e}^{-})^{T}\boldsymbol{\alpha} - 1) - \boldsymbol{\lambda}^{T}\boldsymbol{\alpha} + \boldsymbol{\beta}^{T}(\boldsymbol{\alpha} - \boldsymbol{e})$}
\end{equation}
where $\boldsymbol{\lambda}$ and $\boldsymbol{\beta}$ are Lagrange multiplier vectors corresponding to the inequality constraints 
$\boldsymbol{\alpha} \geq 0 $ and $\boldsymbol{\alpha} \leq 1$, $p$ and $q$ are Lagrange multiplier variables corresponding to the equality constraints $( \boldsymbol{e}^{+})^{T}\mathbf{\boldsymbol{\alpha}} = 1$ and $( \boldsymbol{e}^{-})^{T}\mathbf{\boldsymbol{\alpha}} = 1$, $\boldsymbol{e}^+$ is a column vector where the components corresponding to positive samples are all one and all other components are zero, $\boldsymbol{e}^-$ is a column vector where the components corresponding to negative samples are all one and all other components are zero, and $\boldsymbol{e}$ is a column vector with all components are one. 

L2-SVC-NCH is a convex quadratic programming problem. Thus, $\boldsymbol{\alpha}$ are an optimal solution if and only if there are corresponding Lagrange multiplier variables $p,q,\boldsymbol{\lambda},\boldsymbol{\beta}$ satisfying the KKT condition
\begin{equation}
\label{formula_ex13}
\left\{ \begin{array}{l}
{\nabla f(\boldsymbol{\alpha}) + p\boldsymbol{e}^{+} + q\boldsymbol{e}^{-} - \boldsymbol{\lambda} + \boldsymbol{\beta} = \textbf{0}}, \\
{\boldsymbol{\lambda}^{T}\boldsymbol{\alpha} = 0,~\boldsymbol{\beta}^{T}\left( {\mathbf{\boldsymbol{\alpha}} - \boldsymbol{e}} \right) = 0,\boldsymbol{\lambda} \geq 0,\boldsymbol{\beta} \geq 0}, \\
{( \boldsymbol{e}^{+})^{T}\mathbf{\boldsymbol{\alpha}} = 1,( \boldsymbol{e}^{-})^{T}\mathbf{\boldsymbol{\alpha}} = 1,0 \leq \mathbf{\boldsymbol{\alpha}} \leq 1}. \\
\end{array} \right.
\end{equation}

For positive samples,

if $0 < \boldsymbol{\alpha}_{i}^{+} < 1\Rightarrow\boldsymbol{\lambda}_{i}^{+} = \boldsymbol{\beta}_{i}^{+} = 0\Rightarrow{\nabla}_{i}f( \mathbf{\boldsymbol{\alpha}} ) + p = 0\Rightarrow{- \nabla}_{i}f( \mathbf{\boldsymbol{\alpha}} ) = p,$

if $\boldsymbol{\alpha}_{i}^{+} =0\Rightarrow\boldsymbol{\beta}_{i}^{+} = 0\Rightarrow\nabla_{i}f( \mathbf{\boldsymbol{\alpha}}) + p  = \boldsymbol{\lambda}_{i}^{+} \geq 0 \Rightarrow{ - \nabla}_{i}f( \mathbf{\boldsymbol{\alpha}}) \leq p,$

if $\boldsymbol{\alpha}_{i}^{+} = 1\Rightarrow\boldsymbol{\lambda}_{i}^{+} = 0\Rightarrow{\nabla}_{i}f( \mathbf{\boldsymbol{\alpha}}) + p = {- \boldsymbol{\beta}}_{i}^{+} \leq 0\Rightarrow{ - \nabla}_{i}f( \mathbf{\boldsymbol{\alpha}}) \geq p.$

For negative samples, we have similar results:

if $0 < \boldsymbol{\alpha}_{i}^{-} < 1\Rightarrow\boldsymbol{\lambda}_{i}^{-} = \boldsymbol{\beta}_{i}^{-} = 0\Rightarrow{\nabla}_{i}f( \mathbf{\boldsymbol{\alpha}}) + q = 0\Rightarrow{- \nabla}_{i}f( \mathbf{\boldsymbol{\alpha}}) = q,$

if $\boldsymbol{\alpha}_{i}^{-} =0\Rightarrow\boldsymbol{\beta}_{i}^{-} = 0\Rightarrow\nabla_{i}f( \mathbf{\boldsymbol{\alpha}}) + q  = \boldsymbol{\lambda}_{i}^{-} \geq 0 \Rightarrow{ - \nabla}_{i}f( \mathbf{\boldsymbol{\alpha}}) \leq q,$

if $\boldsymbol{\alpha}_{i}^{-} = 1\Rightarrow\boldsymbol{\lambda}_{i}^{-} = 0\Rightarrow{\nabla}_{i}f( \mathbf{\boldsymbol{\alpha}}) + q = {- \boldsymbol{\beta}}_{i}^{-} \leq 0\Rightarrow{ - \nabla}_{i}f( \mathbf{\boldsymbol{\alpha}}) \geq q.$

Combining the results of positive samples and negative samples, we can obtain \eqref{formula_ex10}. The proof is finished.\#

Let
\begin{align}
I_{up}^{+}&=\left\{ i \middle| {\boldsymbol{\alpha}_{i} < 1~\rm{and}~\boldsymbol{y}_{i} = 1} \right\},\\
{I}_{low}^{+}&=\left\{ i\middle|{\boldsymbol{\alpha}_{i} > 0~\rm{and}~\boldsymbol{y}_{i} = 1}\right\},\\
I_{up}^{-}& = \left\{ i \middle| {\boldsymbol{\alpha}_{i} < 1~\rm{and}~\boldsymbol{y}_{i} = - 1} \right\}, \\
{I}_{low}^{-} &= \left\{ i \middle| {\boldsymbol{\alpha}_{i} > 0~\rm{and}~\boldsymbol{y}_{i} = - 1} \right\}.
\end{align}

From Theorem 1, we have
\begin{align*}
{{{- \nabla}_{i}f(\boldsymbol{\alpha}) \leq p~~\rm{if}~~\textit{i} \in \textit{I}}_{up}^{+},~{{- \nabla}_{i}f(\boldsymbol{\alpha}) \geq p~~\rm{if}~~\textit{i} \in \textit{I}}_{low}^{+}},\\
{{{- \nabla}_{i}f(\boldsymbol{\alpha}) \leq q~~\rm{if}~~\textit{i} \in \textit{I}}_{up}^{-},~{{- \nabla}_{i}f(\boldsymbol{\alpha}) \geq q~~\rm{if}~~\textit{i} \in \textit{I}}_{low}^{-}}.
\end{align*}
$\Rightarrow$
\begin{equation}
\label{formula_ex18}
\textit{m}^{+}(\boldsymbol{\alpha}) \leq M^{+}(\boldsymbol{\alpha})~\rm{and}~\textit{m}^{-}(\boldsymbol{\alpha}) \leq \textit{M}^{-}(\boldsymbol{\alpha})
\end{equation}

where

\begin{equation}
\label{formula_ex19}
\begin{matrix}
{\textit{m}^{+}(\boldsymbol{\alpha}) = {\max\limits_{i \in I_{\mathit{up}}^{+}}{{- \nabla}_{i}f(\boldsymbol{\alpha})}},M^{+}(\boldsymbol{\alpha}) = {\min\limits_{i \in I_{\mathit{low}}^{+}}{{- \nabla}_{i}f(\boldsymbol{\alpha})}}}, \\
{\textit{m}^{-}(\boldsymbol{\alpha}) = {\max\limits_{i \in I_{\mathit{up}}^{-}}{{- \nabla}_{i}f(\boldsymbol{\alpha})}},M^{-}(\boldsymbol{\alpha}) = {\min\limits_{i \in I_{\mathit{low}}^{-}}{{- \nabla}_{i}f(\boldsymbol{\alpha})}}}. \\
\end{matrix}
\end{equation}
\\

\textbf{Corollary 1} Another KKT condition of the optimal solution $\boldsymbol{\alpha}$ for L2-SVC-NCH is $\textit{m}^{+}(\boldsymbol{\alpha}) \leq M^{+}(\boldsymbol{\alpha})$ and $\textit{m}^{-}(\boldsymbol{\alpha}) \leq M^{-}(\boldsymbol{\alpha})$.

The KKT condition \eqref{formula_ex10} with a given precision $\varepsilon$ can be descripted as

\begin{equation}
\label{formula_ex20}
\begin{array}{l}
{\forall~0 < \boldsymbol{\alpha}_{i}^{+} < 1,~\left| {{- \nabla}_{i}f\left( \mathbf{\boldsymbol{\alpha}} \right) - \mu^{+}} \right| \leq \varepsilon,} \\
{\forall~\boldsymbol{\alpha}_{i}^{+} = 0,~- \nabla_{i}f\left( \mathbf{\boldsymbol{\alpha}} \right) \leq \mu^{+} + \varepsilon,} \\
{\forall~\boldsymbol{\alpha}_{i}^{+} = 1,~- \nabla_{i}f\left( \mathbf{\boldsymbol{\alpha}} \right) \geq \mu^{+} - \varepsilon,} \\
\left. \forall~0 < \boldsymbol{\alpha}_{i}^{-} < 1,~\middle| {- \nabla}_{i}f\left( \mathbf{\boldsymbol{\alpha}} \right) - \mu^{-} \middle| \leq \varepsilon, \right. \\
{\forall~\boldsymbol{\alpha}_{i}^{-} = 0,~- \nabla_{i}f\left( \mathbf{\boldsymbol{\alpha}} \right) \leq \mu^{-} + \varepsilon,} \\
{\forall~\boldsymbol{\alpha}_{i}^{-} = 1,~- \nabla_{i}f\left( \mathbf{\boldsymbol{\alpha}} \right) \geq \mu^{-} - \varepsilon} 
\end{array}
\end{equation}where
\begin{equation}
\label{formula_ex21}
\begin{matrix}
{{\mu^{+} = {\sum\limits_{0 < \alpha_{i}^{+} < 1}{\left( - \nabla \right._{i}f(\alpha)}})/{\overset{-}{l}}_{+}},} \\
{{\mu^{-} = {\sum\limits_{0 < \alpha_{i}^{-} < 1}{\left( - \nabla \right._{i}f(\alpha)}})/{\overset{-}{l}}_{-}},} \\
\end{matrix}
\end{equation}
${\overset{-}{l}_{+}}$ and ${\overset{-}{l}_{-}}$ are the number of samples that satisfy $0 < \boldsymbol{\alpha}_{i}^{+} < 1$ and $0 < \boldsymbol{\alpha}_{i}^{-} < 1$ respectively. The KKT condition \eqref{formula_ex20} can be used in the corresponding algorithm.

Similarly, the KKT condition \eqref{formula_ex18}  with a given precision can be descripted as
\begin{equation}
\label{formula_ex22}
\textit{m}^{+}(\boldsymbol{\alpha}) \leq M^{+}(\boldsymbol{\alpha}) + \varepsilon~{\rm and}~
{\textit{m}}^{-}(\boldsymbol{\alpha})\leq {M^{-}(\boldsymbol{\alpha})} + \varepsilon.
\end{equation}
The KKT condition \eqref{formula_ex22} can be used in the sequential minimal optimization (SMO) algorithm based on maximal violation pair.\\

\subsubsection{A projected gradient algorithm}
The most commonly used algorithms to solve the minimization problem L2-SVC-NCH in \eqref{formula_ex7} are SMO algorithms \cite{1998Sequential,keerthi2001improvements,fan2005working}. L2-SVC-NCH is a strictly convex quadratic programming problem, which can also be solved by PGA \cite{2005optimization,Linear}. The PGA is flexible due to its simplicity. In fact, the SMO is a special case of the PGA. Therefore, we provide a PGA to solve L2-SVC-NCH.

In PGA, a feasible descent direction is first obtained by the projection of negative gradient vector with a projection matrix. Then, one-dimensional search in the feasible descent direction is performed so that the objective function is decreased.

The constraint $0 \leq \boldsymbol{\alpha}_{i} \leq 1$ can be expressed as $0 \leq \left. \left( \boldsymbol{e} \right._{i} \right)^{T}\boldsymbol{\alpha} \leq 1$ where $e_i$ is a column vector whose the i-th component is one and other components are zero. Set
\begin{equation}
\label{formula_ex23}
\boldsymbol{M} = \begin{bmatrix}\boldsymbol{E} \\( \boldsymbol{e} ^{+} )^{T} \\( \boldsymbol{e}^{-} )^{T} \\
\end{bmatrix}_{{(l}_{0} + 2) \times l}
\end{equation}
where $l_0$ is the number of the effective constraints $\left.\{ \left.\left( \boldsymbol{e} \right._{i} \right)^{T}\boldsymbol{\alpha} = 0~\mathrm{or}~\left.
\left( \boldsymbol{e} \right._{i} \right)^{T}\boldsymbol{\alpha} = 1\right.$\\
$\left.|i = 1,2,\cdots,l \right.\}$, ~$\boldsymbol{E} = \left( \left. \left( \boldsymbol{e} \right._{i} \right)^{T} \right)_{l_{0} \times l}$, $\left( \boldsymbol{e}^{+} \right)^{T}$ and $\left( \boldsymbol{e}^{-} \right)^{T}$ are the coefficient vectors corresponding to the equality constrains $( \boldsymbol{e} ^{+})^{T}\boldsymbol{\alpha} = 1$ and $(\textbf{e}^{-} )^{T}\boldsymbol{\alpha} = 1$ respectively. The projection matrix
\begin{equation}
\label{formula_ex24}
\boldsymbol{P} = \boldsymbol{I} - \boldsymbol{M}^{T}\left( {\boldsymbol{MM}^{T}} \right)^{- 1}\boldsymbol{M}.
\end{equation}
The projection of negative gradient vector with $\boldsymbol{P}$ is
\begin{equation}
\label{formula_ex25}
\boldsymbol{d} = - \boldsymbol{P}\nabla f(\boldsymbol{\alpha})
\end{equation}
where
\begin{equation}
\label{formula_ex26}
\boldsymbol{d}_{i} = \begin{cases}
\begin{aligned}
  &~~~~~~~~0,\quad &&\text{if}~\boldsymbol{\alpha}_{i} = 0~\text{or}~\boldsymbol{\alpha}_{i} = 1, \\
  &{- \nabla}_{i}f(\boldsymbol{\alpha}) - \mu^{+}, \quad &&\text{if}~0 < \boldsymbol{\alpha}_{i} < 1~\text{and}~\boldsymbol{y}_{i} = 1, \quad i = 1,2,\ldots,l \\
  &{- \nabla}_{i}f(\boldsymbol{\alpha}) - \mu^{-}, \quad &&\text{if}~0 < \boldsymbol{\alpha}_{i} < 1~\text{and}~\boldsymbol{y}_{i} = -1.
\end{aligned}
\end{cases}
\end{equation}The corresponding multiplier variables for the effective constraints and the equality constraints are
\begin{equation}
\label{formula_ex27}
\boldsymbol{\mu} = {- \left( {\boldsymbol{MM}^{T}} \right)}^{- 1}\boldsymbol{M}\nabla f(\boldsymbol{\alpha})
\end{equation}
where
\begin{equation}
\label{formula_ex28}
\boldsymbol{\mu}_{i} = \begin{cases}
\begin{aligned}
  &~~~\nabla_{i}f(\boldsymbol{\alpha}) + \mu^{+}, &&\text{for}~\left( \boldsymbol{e}_{i}\right)^{T}\boldsymbol{\alpha} = 0~\text{and}~\boldsymbol{y}_{i} = 1, \\
  &~~~\nabla_{i}f(\boldsymbol{\alpha}) + \mu^{-}, &&\text{for}~\left( \boldsymbol{e}_{i}\right)^{T}\boldsymbol{\alpha} = 0~\text{and}~\boldsymbol{y}_{i} = -1, \\
  &-\nabla_{i}f(\boldsymbol{\alpha}) - \mu^{+}, &&\text{for}~\left( \boldsymbol{e}_{i}\right)^{T}\boldsymbol{\alpha} = 1~\text{and}~\boldsymbol{y}_{i} = 1, \quad ~i = 1,2,\ldots,l_{0} + 2\\
  &-\nabla_{i}f(\boldsymbol{\alpha}) - \mu^{-}, &&\text{for}~\left( \boldsymbol{e}_{i}\right)^{T}\boldsymbol{\alpha} = 1~\text{and}~\boldsymbol{y}_{i} = -1, \\
  &~~~~~~~~~~\mu^{+}, &&\text{for}~( \boldsymbol{e}^{+})^{T}\boldsymbol{\alpha} = 1, \\
  &~~~~~~~~~~\mu^{-}, &&\text{for}~( \boldsymbol{e}^{-})^{T}\boldsymbol{\alpha} = 1.
\end{aligned}
\end{cases}
\end{equation}

If $\boldsymbol{d} \neq \textbf{0}$, then $\boldsymbol{d}$ is a feasible descent direction. If $\boldsymbol{d}=\textbf{0}$ and $\boldsymbol{\mu}_{i} \geq 0$ for $\left. \left( \boldsymbol{e} \right._{i} \right)^{T}\boldsymbol{\alpha} = 0$ or $\left. \left( \boldsymbol{e} \right._{i} \right)^{T}\boldsymbol{\alpha} = 1$, then $\boldsymbol{\alpha}$ is an optimal solution. If $\boldsymbol{d}=\textbf{0}$ and there is $\boldsymbol{\mu}_{i} < 0$ for $\left. \left( \boldsymbol{e} \right._{i} \right)^{T}\boldsymbol{\alpha} = \textbf{0}$ or $\left. \left( \boldsymbol{e} \right._{i} \right)^{T}\boldsymbol{\alpha} = 1$, then a non-zero feasible descent direction can be obtained by removing the corresponding row from $\boldsymbol{M}$ and re-computing the projection matrix $\boldsymbol{P}$ with the new $\boldsymbol{M}$. 

It should be noted that we don't need to compute $ (MM^T)^{-1} $ in $ d $ and $ \mu $. In fact, they can be deduced analytically, and we just need to calculate them according to \eqref{formula_ex26} and \eqref{formula_ex28} where computing $ (MM^T)^{-1} $ is not needed. After a non-zero feasible descent direction $\boldsymbol{d}$ is obtained, the one-dimensional search is modeled as
\begin{equation}
\label{formula_ex29}
\underset{\mathit{\eta~}}{\min~}{\frac{1}{2}\left( \left. \boldsymbol{\alpha} + \eta \boldsymbol{d} \right)^{T}(\boldsymbol{G} + {\boldsymbol{I}}/C)(\boldsymbol{\alpha} + \eta \boldsymbol{d}) \right.}
\end{equation}
\begin{center}
$s.t.~~0 \leq \eta \leq \eta_{max}$
\end{center}
i.e.,
\begin{equation}
\label{formula_ex30}
\underset{\mathit{\eta~}}{\min~}{\frac{1}{2}\boldsymbol{d}^{T}(\boldsymbol{G} + \boldsymbol{I}/C)\boldsymbol{d}\eta^{2} + \boldsymbol{d}^{T}(\boldsymbol{G} + \boldsymbol{I}/C)\boldsymbol{\alpha}\eta}
\end{equation}
\begin{center}
$s.t.~~0 \leq \eta \leq \eta_{max}$
\end{center}
where
\begin{equation}
\label{formula_ex31}
\eta_{max} = \rm{min}\left\{ \eta_{max0}^{+},~\eta_{max1}^{+},~\eta_{max0}^{-},~\eta_{max1}^{-}\right\},
\end{equation}

\begin{equation}
\label{formula_ex32}
\left\{\begin{array}{l}
\eta_{max0}^{+} = {\min\limits_{i:0 < \boldsymbol{\alpha}_{i} < 1,y_{i} =  1,\nabla_{i}f{(\mathbf{\boldsymbol{\alpha}})} + \mu^{+} > 0}{\boldsymbol{\alpha}_{i}/\left( \nabla_{i}f\left( \mathbf{\boldsymbol{\alpha}} \right) + \mu^{+} \right)}},\\

\eta_{max1}^{+} = {\min\limits_{i:0 < \boldsymbol{\alpha}_{i} < 1,y_{i} =  1,\nabla_{i}f{(\mathbf{\boldsymbol{\alpha}})} + \mu^{+} < 0}{(1-{\boldsymbol{\alpha}_{i})/\left( -\nabla_{i}f\left( \mathbf{\boldsymbol{\alpha}} \right) - \mu^{+} \right)}.}}\end{array} \right.
\end{equation}

\begin{equation}
\label{formula_ex33}
\left\{\begin{array}{l}
\eta_{max0}^{-} = {\min\limits_{i:0 < \boldsymbol{\alpha}_{i} < 1,y_{i} = - 1,\nabla_{i}f{(\mathbf{\boldsymbol{\alpha}})} + \mu^{-} > 0}{\boldsymbol{\alpha}_{i}/\left( \nabla_{i}f\left( \mathbf{\boldsymbol{\alpha}} \right) + \mu^{-} \right)}},\\

\eta_{max1}^{-} = {\min\limits_{i:0 < \boldsymbol{\alpha}_{i} < 1,y_{i} = - 1,\nabla_{i}f{(\mathbf{\boldsymbol{\alpha}})} + \mu^{-} < 0}{(1-{\boldsymbol{\alpha}_{i})/\left( -\nabla_{i}f\left( \mathbf{\boldsymbol{\alpha}} \right) - \mu^{-} \right)}.}}\end{array} \right.
\end{equation}

Set
\begin{equation}
\label{formula_ex34}
g(\eta) = \frac{1}{2}\boldsymbol{d}^{T}(\boldsymbol{G} + \boldsymbol{I}/C)\boldsymbol{d}\eta^{2} + \boldsymbol{d}^{T}(\boldsymbol{G} + \boldsymbol{I}/C)\boldsymbol{\alpha}\eta.
\end{equation}
Let $g'(\eta) = 0$, we obtain
\begin{equation}
\label{formula_ex35}
\overset{-}{\eta} = - \frac{\boldsymbol{d}^{T}(\boldsymbol{G} + \boldsymbol{I}/C)\boldsymbol{\alpha}}{\boldsymbol{d}^{T}(\boldsymbol{G} + \boldsymbol{I}/C)\boldsymbol{d}}.
\end{equation}
Thus, the optimal solution of \eqref{formula_ex30} is
\begin{equation}
\label{formula_ex36}
\eta^{*} = \left\{ \begin{matrix}
{\overset{-}{\eta},~\rm{if}~0 \leq \overset{-}{\eta} \leq \eta_{max}}, \\
{~~~0,~~~\rm{if}~\overset{-}{\eta} < 0,~~~~} \\
{\eta_{max},~\rm{if}~\overset{-}{\eta} > \eta_{max}.} \\
\end{matrix} \right.
\end{equation}
and $\boldsymbol{\alpha}$ is updated according to
\begin{equation}
\label{formula_ex37}
\boldsymbol{\alpha} = \boldsymbol{\alpha} + \eta^{*}\boldsymbol{d}.
\end{equation}

A projected gradient algorithm for L2-SVC-NCH is provided in Algorithm \ref{alg:alg1} where $k_{type}$ and $k_{para}$ are the type and the parameter of kernel function. 
\begin{algorithm}
\caption{$\boldsymbol{\alpha}^{*}=PGA(T, k_{type}, k_{para}, C, \varepsilon, epoch)$}
\label{alg:alg1}
\begin{algorithmic}[1]
\State \textbf{Input} $T=\left\{( \boldsymbol{x}_{i},\boldsymbol{y}_{i} \right)\left|{\boldsymbol{x}_{i} \in \textbf{R}^{n},\boldsymbol{y}_{i}\in \left\{ {+ 1, - 1} \right\},i = 1,\ldots,l} \right\}$,$k_{type}, k_{para},C,$\\ $\varepsilon,epoch$
\State \textbf{Output} $\boldsymbol{\alpha}^*$
\State \textbf{Initialization} $\mathbf{\boldsymbol{\alpha}}_{i \in {ID}^{+}} = \frac{1}{l^{+}}~,~~\mathbf{\boldsymbol{\alpha}}_{i \in {ID}^{-}} = \frac{1}{l^{-}}$,\\$l^{+}$ and $l^{-}$ are the number of positive and negative samples
\State Compute $\boldsymbol{G}$ by $k_{type}$ and $k_{para}$
\State \textbf{For} $k=1:epoch$
\State \hspace{0.25cm}Compute $\nabla f(\boldsymbol{\alpha})$ by \eqref{formula_ex11}
\State \hspace{0.25cm}Compute $\mu^{+}$ and $\mu^{-}$ by \eqref{formula_ex21}
\State \hspace{0.25cm}Compute $\boldsymbol{d}$ by \eqref{formula_ex26}
\State \hspace{0.25cm}\textbf{If} $\left| \middle| \boldsymbol{d} \middle| \right|_{\infty} \leq \varepsilon$
\State \hspace{0.5cm}Compute $\boldsymbol{\mu}$ by \eqref{formula_ex28}
\State \hspace{0.5cm}\textbf{If} exists $\boldsymbol{\mu}_{i} < 0$ for $\left. \left( \boldsymbol{e} \right._{i} \right)^{T}\boldsymbol{\alpha} = 0$
\State \hspace{0.75cm}$j \gets {\arg{\rm{min}\left\{ \boldsymbol{\mu}_{i} < 0~for~\left. \left( \boldsymbol{e} \right._{i} \right)^{T}\boldsymbol{\alpha} = 0 \right\}}}$
\State \hspace{0.75cm}$\boldsymbol{d}_{j} \gets 0$
\State \hspace{0.75cm}\textbf{If} $\boldsymbol{y}_{j} = 1$
\State \hspace{1cm}$\mu^{+} \gets {\sum\limits_{0 < \boldsymbol{\alpha}_{i}^{+} < 1,i \neq j}{ - \nabla_{i}f( \mathbf{\boldsymbol{\alpha}})}} /( {\overset{-}{l}}_{+} - 1)$
\State \hspace{1cm}$d_{i} \gets {- \nabla}_{i}f(\boldsymbol{\alpha}) - \mu^{+}$ for $\boldsymbol{\alpha}_{i} > 0$ and $i \neq j$
\State \hspace{0.75cm}\textbf{Else}
\State \hspace{1cm}$\mu^{-} \gets {\sum\limits_{0 < \boldsymbol{\alpha}_{i}^{-} < 1,i \neq j}{ - \nabla_{i}f( \mathbf{\boldsymbol{\alpha}})}} /( {\overset{-}{l}}_{-} - 1)$
\State \hspace{1cm}$\boldsymbol{d}_{i} \gets {- \nabla}_{i}f(\boldsymbol{\alpha}) - \mu^{-}$ for $\boldsymbol{\alpha}_{i} > 0$ and $i \neq j$
\State \hspace{0.75cm}\textbf{End If}
\State \hspace{0.5cm}\textbf{Else}
\State \hspace{0.75cm}\textbf{Break}
\State \hspace{0.5cm}\textbf{End If}
\State \hspace{0.25cm}\textbf{End If}
\State \hspace{0.25cm}Compute $\eta^*$ by \eqref{formula_ex31} to \eqref{formula_ex36}
\State \hspace{0.25cm}$\mathbf{\boldsymbol{\alpha}} \gets \mathbf{\boldsymbol{\alpha}} + \eta^{*}\boldsymbol{d}$
\State \textbf{End For}
\State $\mathbf{\boldsymbol{\alpha}^*}\gets \mathbf{\boldsymbol{\alpha}}$
\State \textbf{Return}
\end{algorithmic}
\label{alg1}
\end{algorithm}

\subsubsection{A SMO algorithm}
To carry out the comparison of the PGA and SMO algorithms, we also provide a SMO algorithm of L2-SVC-NCH. SMO algorithms generally consist of two main steps: finding two variables by some heuristic rules and solving the optimization problem formed by the two variables. 

For L2-SVC-NCH, the two variables can be obtained with maximal violating pair. A maximal violating pair refers to a pair of samples with the most violation to KKT condition \eqref{formula_ex18}. If the KKT condition \eqref{formula_ex18} is not satisfied, and
\begin{equation}
\label{formula_ex38}
\begin{matrix}
{i^{+} = \mathrm {arg}~{\max\limits_{i \in I_{\mathit{up}}^{+}}{{- \nabla}_{i}f(\boldsymbol{\alpha})}},j^{+} = \mathrm {arg}{\min\limits_{i \in I_{\mathit{low}}^{+}}{{- \nabla}_{i}f(\boldsymbol{\alpha})}}}, \\
{i^{-} = \mathrm {arg}~{\max\limits_{i \in I_{\mathit{up}}^{-}}{{- \nabla}_{i}f(\boldsymbol{\alpha})}},j^{-} = \mathrm {arg}{\min\limits_{i \in I_{\mathit{low}}^{-}}{{- \nabla}_{i}f(\boldsymbol{\alpha})}}}, \\
\end{matrix}
\end{equation}
then the most violating pair is
\begin{equation}
\label{formula_ex39}
\text{$\left( {i,j} \right) = \mathrm {arg}{\max\limits_{{({i^{+},j^{+}})},{({i^{-},j^{-}})}}\left\{ m^{+}\left( \boldsymbol{\alpha}_{i^{+}} \right) - M^{+}\left( \boldsymbol{\alpha}_{j^{+}} \right),~ \right.}m^{-}\left( \boldsymbol{\alpha}_{i^{-}} \right) - M^{-}\left( \boldsymbol{\alpha}_{j^{-}} \right)\}$}.
\end{equation}

After the most violation pair $\left( {i,j} \right)$ is obtained, the optimization problem formed by the two variables is

\begin{equation}
\label{formula_ex40}
\max_{\boldsymbol{\alpha}_{i},\boldsymbol{\alpha}_{j}} \frac{1}{2}\left( k(\boldsymbol{x}_{i},\boldsymbol{x}_{i}) + C \right)\boldsymbol{\alpha}_{i}^{2} + \frac{1}{2}\left( k(\boldsymbol{x}_{j},\boldsymbol{x}_{j})+C \right)\boldsymbol{\alpha}_{j}^{2} +\boldsymbol{y}_{i}\boldsymbol{y}_{j}k(\boldsymbol{x}_{i},\boldsymbol{x}_{j})\boldsymbol{\alpha}_{i}\boldsymbol{\alpha}_{j} + \boldsymbol{v}_{i}\boldsymbol{\alpha}_{i} + \boldsymbol{v}_{j}\boldsymbol{\alpha}_{j} \notag
\end{equation}

\begin{equation}
% \begin{center}
s.t.~~\boldsymbol{\alpha}_{i} + \boldsymbol{\alpha}_{j} = \boldsymbol{\alpha}_{i}^{old} + \boldsymbol{\alpha}_{j}^{old} = Const,~0 \leq \boldsymbol{\alpha}_{i},\boldsymbol{\alpha}_{j} \leq 1
% \end{center}
\end{equation}

where
\begin{equation}
\label{formula_ex41}
\boldsymbol{v}_{i} = \boldsymbol{y}_{i}{\sum\limits_{k \neq i,j}{\boldsymbol{y}_{k}k\left( \boldsymbol{x}_{i},\boldsymbol{x}_{k} \right)\boldsymbol{\alpha}_{k}}},\boldsymbol{v}_{j} = \boldsymbol{y}_{j}{\sum\limits_{k \neq i,j}{\boldsymbol{y}_{k}k\left( \boldsymbol{x}_{j},\boldsymbol{x}_{k} \right)\boldsymbol{\alpha}_{k}}}.
\end{equation}

Suppose $\left( \boldsymbol{\alpha}_{i}^{uc},\boldsymbol{\alpha}_{j}^{uc} \right)$ is the optimal solution of \eqref{formula_ex40} without considering the constraint $0 \leq \boldsymbol{\alpha}_{i},\boldsymbol{\alpha}_{j} \leq 1$, then \\
\begin{equation}
\label{formula_ex42_1}
\boldsymbol{\alpha}_{i}^{uc} =\frac{\left( {k\left( {\boldsymbol{x}_{j},\boldsymbol{x}_{j}} \right) + C} \right)\left( {\boldsymbol{\alpha}_{i}^{old} + \boldsymbol{\alpha}_{j}^{old}} \right) - \boldsymbol{y}_{i}\boldsymbol{y}_{j}k\left( {\boldsymbol{x}_{i},\boldsymbol{x}_{j}} \right)\left( {\boldsymbol{\alpha}_{i}^{old} + \boldsymbol{\alpha}_{j}^{old}} \right) - \boldsymbol{v}_{i} + \boldsymbol{v}_{j}}{k\left( {\boldsymbol{x}_{i},\boldsymbol{x}_{i}} \right) + k\left( {\boldsymbol{x}_{j},\boldsymbol{x}_{j}} \right) + 2C - 2\boldsymbol{y}_{i}\boldsymbol{y}_{j}k\left( {\boldsymbol{x}_{i},\boldsymbol{x}_{j}} \right)}
\end{equation}
and
\begin{equation}
\label{formula_ex42}
\boldsymbol{\alpha}_{j}^{uc} = \boldsymbol{\alpha}_{i}^{old} + \boldsymbol{\alpha}_{j}^{old} - \boldsymbol{\alpha}_{i}^{uc}.
\end{equation}
The optimal solution of \eqref{formula_ex40}  with the constraint $0 \leq \boldsymbol{\alpha}_{i},\boldsymbol{\alpha}_{j} \leq 1$ is
\begin{equation}
\label{formula_ex43}
\left. \left( \boldsymbol{\alpha} \right._{i}^{*},\boldsymbol{\alpha}_{j}^{*} \right) = \left\{ \begin{matrix}
{\left. {}\left( \boldsymbol{\alpha} \right._{i}^{uc},\boldsymbol{\alpha}_{j}^{uc} \right),~if~0 \leq \boldsymbol{\alpha}_{i}^{uc} \leq \boldsymbol{\alpha}_{i}^{old} + \boldsymbol{\alpha}_{j}^{old},~~} \\
{\left( {0,~\boldsymbol{\alpha}_{i}^{old} + \boldsymbol{\alpha}_{j}^{old}} \right),~~~~~if~\boldsymbol{\alpha}_{i}^{uc} < 0,~~~~~~~~~~} \\
{( \boldsymbol{\alpha}_{i}^{old} + \boldsymbol{\alpha}_{j}^{old},0 ),~if~\boldsymbol{\alpha}_{i}^{uc} > \boldsymbol{\alpha}_{i}^{old} + \boldsymbol{\alpha}_{j}^{old}.~~} \\
\end{matrix} \right.
\end{equation}

A SMO algorithm for L2-SVC-NCH is provided in Algorithm \ref{alg:alg2}.
\begin{algorithm}
\caption{$\boldsymbol{\alpha}^{*} = SMO(T, k_{type}, k_{para}, C,\varepsilon,epoch)$}
\label{alg:alg2}
\begin{algorithmic}[1] 
\State \textbf{Input} $T=\left\{( \boldsymbol{x}_{i},\boldsymbol{y}_{i} \right)\left|{\boldsymbol{x}_{i} \in \textbf{R}^{n},\boldsymbol{y}_{i}\in \left\{ {+ 1, - 1} \right\},i = 1,\ldots,l} \right\}$,$k_{type}, k_{para},C,$\\
$\varepsilon,epoch$
\State \textbf{Output} $\boldsymbol{\alpha}^*$
\State \textbf{Initialization} $\mathbf{\boldsymbol{\alpha}}_{i \in {ID}^{+}} = \frac{1}{l^{+}}~,~~\mathbf{\boldsymbol{\alpha}}_{i \in {ID}^{-}} = \frac{1}{l^{-}}$
\State Compute $\boldsymbol{G}$ by $k_{type}$ and $k_{para}$
\State \textbf{For} $k=1:epoch$
\State \hspace{0.25cm}Compute $\nabla f(\boldsymbol{\alpha})$ by \eqref{formula_ex11}
\State \hspace{0.25cm}Compute $i^{+},j^{+},i^{-},j^{-}$ by \eqref{formula_ex38}
\State \hspace{0.25cm}\textbf{If} $m^{+}\left( \boldsymbol{\alpha}_{i^{+}} \right) \leq M^{+}\left( \boldsymbol{\alpha}_{j^{+}} \right) + \varepsilon$ and $m^{-}\left( \boldsymbol{\alpha}_{i^{-}} \right) \leq M^{-}\left( \boldsymbol{\alpha}_{j^{-}} \right) + \varepsilon$
\State \hspace{0.5cm}\textbf{Break}
\State \hspace{0.25cm}\textbf{End If}
\State \hspace{0.25cm}Find the most violating pair $(i,j)$ by \eqref{formula_ex39}
\State \hspace{0.25cm}Compute $\boldsymbol{v}_i,\boldsymbol{v}_j$ by \eqref{formula_ex41}
\State \hspace{0.25cm}Compute $\boldsymbol{\alpha}_i^{uc},\boldsymbol{\alpha}_j^{uc}$  by \eqref{formula_ex42_1} and \eqref{formula_ex42}
\State \hspace{0.25cm}\textbf{Update} $\boldsymbol{\alpha}_i,\boldsymbol{\alpha}_j$ by \eqref{formula_ex43}
\State \textbf{End For}
\State $\mathbf{\boldsymbol{\alpha}^*}\gets \mathbf{\boldsymbol{\alpha}}$
\State \textbf{Return}
\end{algorithmic}
\label{alg2}
\end{algorithm}

\subsubsection{A comparison of the PGA and the SMO algorithm}
The space complexity of the PGA and the SMO algorithm for solving L2-SVC-NCH are the same order, and their time complexity are also close. The variables in the PGA and the SMO that need to be stored are $\boldsymbol{G},\boldsymbol{\alpha},\nabla f(\boldsymbol{\alpha})$, which are the same. The number of multiplication in one iteration of the PGA is $O( 2l ^{2} + 2l )$ while it is $O( l ^{2} + 4l )$ for the SMO. Thus, the number of multiplication of the PGA and the SMO algorithm are $O\left( {\left( {{2l}^{2} + 2l} \right) \times m_{p}} \right)$ and $O\left( {\left( {{l}^{2} + 4l} \right) \times m_{s}} \right)$ where $m_p$  and $m_s$ are the number of iteration for the PGA and the SMO. Considering that both require $O(l ^{2} )$ , their time complexity are close.

Further, the SMO algorithm can be regarded as a special case of the PGA. In the original PGA, almost all components of $\boldsymbol{\alpha}$ (except $\boldsymbol{\alpha}_{i} = 0$) are updated in one iteration. However, the PGA can also only update two components as the SMO does. From \eqref{formula_ex26}, we can select the two components with the largest standard deviation if we need to select two components among positive samples to be updated, i.e.
\begin{equation}
\label{formula_ex44}
\left( {i^{+},j^{+}} \right) = {\arg{\max\limits_{\boldsymbol{y}_{i} = \boldsymbol{y}_{j} = 1}{\mathrm{std}\left\{ {- \nabla}_{i}f(\boldsymbol{\alpha}),{- \nabla}_{j}f(\boldsymbol{\alpha}) \right\}}}}.
\end{equation}
If $(i^+,j^+ )$  is satisfied \eqref{formula_ex44}, then $\boldsymbol{d}_{i^{+}}$  and  $\boldsymbol{d}_{j^{+}}$   are large. Thus $\boldsymbol{\alpha}_{i^{+}}$ and $\boldsymbol{\alpha}_{j^{+}}$ can be updated with large values. The $\left( {i^{+},j^{+}} \right)$  that is satisfied \eqref{formula_ex44} can also be obtained by
\begin{equation}
\label{formula_ex45}
i^{+} = {\arg{\max\limits_{\boldsymbol{y}_{i} = 1}{{- \nabla}_{i}f(\boldsymbol{\alpha})}}},~~j^{+} = {\arg{\min\limits_{\boldsymbol{y}_{j} = 1}{{- \nabla}_{j}f(\boldsymbol{\alpha})}}}.
\end{equation}
If we select two components among negative samples to be updated, then
\begin{equation}
\label{formula_ex46}
i^{-} = {\arg{\max\limits_{\boldsymbol{y}_{i} = - 1}{{- \nabla}_{i}f(\boldsymbol{\alpha})}}},~~j^{-} = {\arg{\min\limits_{\boldsymbol{y}_{j} = - 1}{{- \nabla}_{j}f(\boldsymbol{\alpha})}}}.
\end{equation}

We see that the two variables selected by the PGA and the SMO are same by the comparison of \eqref{formula_ex38} and \eqref{formula_ex45}, \eqref{formula_ex46}. In addition, both the solution obtained from \eqref{formula_ex37} and the one from \eqref{formula_ex43} are the optimal solution of the two-variables optimal problem. Thus, the solutions of the PGA and the SMO are same since the optimal solution is unique. Hence, the SMO is a special case of the PGA in which only two components of the projected gradient vector are non-zero. In fact, the number of the components updated in each iteration by the PGA can vary from two to all. Hence, the PGA is more flexible compared to the SMO in solving L2-SVC-NCH, i.e. the minimization problem of \eqref{formula_ex7}.

\subsection{The solution of the maximization problem}

The maximization problem in \eqref{formula_ex7} is a univariate optimization problem. We provide a GA-DLR algorithm to solve it. Suppose $\boldsymbol{\alpha}^{*}$ is an optimal solution of L2-SVC-NCH, then the derivative for the maximization problem with the given $\boldsymbol{\alpha}^{*}$  is
\begin{equation}
\label{formula_ex48}
f^{'}(\gamma) = \frac{1}{2}\left. \left( \boldsymbol{\alpha} \right.^{*} \right)^{T}\left\lbrack {\boldsymbol{y}_{i}\boldsymbol{y}_{j}{\left( {- \left\| {\boldsymbol{x}_{i} - \boldsymbol{x}_{j}} \right\|^{2}} \right)e}^{- \gamma{\|{\boldsymbol{x}_{i} - \boldsymbol{x}_{j}}\|}^{2}}} \right\rbrack_{l \times l}\boldsymbol{\alpha}^{\mathbf{*}}.
\end{equation}
A gradient-based rule for updating $\gamma$ is
\begin{equation}
\label{formula_ex49}
\gamma = \gamma + \eta f^{'}(\gamma)
\end{equation}
where $\eta$ is the learning rate.

We provide a strategy of dynamic learning rate for the selection of $\eta$ so that the iteration of $\gamma$ can be speeded up. The principle of the strategy is as follows: $\eta$ will take a relatively large value if the latest update for $\gamma$ is large, otherwise $\eta$ will take a small value because $\gamma$ may be near the extreme point. The provided strategy is
\begin{equation}
\label{formula_ex50}
\eta = \left| {\gamma^{\mathbf{n}\mathbf{e}\mathbf{w}} - \gamma^{\mathbf{o}\mathbf{l}\mathbf{d}}} \right|
\end{equation}
and the initial value of $\eta$ is set with 1. In addition, $\eta$ needs to satisfy that $
f(\gamma)$ is ascending. If $f(\gamma)$ is not ascending, $\eta$ will be halved until $
f(\gamma)$ is ascending.
\subsection{The solution of the minimax problem}
Through connecting the PGA and the GA-DLR in series, a gradient-based (GB) algorithm is obtained for the solution of MaxMin-L2-SVC-NCH. The following Algorithm 3 provides the GB algorithm where $\boldsymbol{\alpha}$ and $\gamma$ are optimized alternately. If the stop condition about gradient with the given precision is satisfied in Algorithm 3, $\left( {\boldsymbol{\alpha}^{*},\gamma^{*}} \right)$ is a local suboptimal solution of \eqref{formula_ex7}.

\begin{algorithm}
\caption{$\left\{ \boldsymbol{\alpha}^{*},\gamma^{*} \right\}=GB\left(T,'Gauss',\gamma_{0},C_{0},\varepsilon_{1},\varepsilon_{2},epoch_\gamma, epoch_\alpha \right)$}
\label{alg:alg3}
\begin{algorithmic}[1]
\State \textbf{Input} $T=\left\{( \boldsymbol{x}_{i},\boldsymbol{y}_{i} \right)\left|{\boldsymbol{x}_{i} \in \textbf{R}^{n},\boldsymbol{y}_{i}\in \left\{ {+ 1, - 1} \right\},i = 1,\ldots,l} \right\}$,$'Gauss',\gamma_{0},C_{0}$\\$\varepsilon_{1},\varepsilon_{2},epoch_\gamma, epoch_\alpha$
\State \textbf{Output} $\left\{ \boldsymbol{\alpha}^{*},\gamma^{*} \right\}$
\State \textbf{Initialization} $\mathbf{\boldsymbol{\alpha}}_{i \in {ID}^{+}} = \frac{1}{l^{+}}, \mathbf{\boldsymbol{\alpha}}_{i\in {ID}^{-}} = \frac{1}{l^{-}},\gamma = \gamma_{0},C = C_{0}$
\State \textbf{For} $n = 1:~epoch_\gamma$
\State \hspace{0.25cm}$\boldsymbol{\alpha}^{*} = PGA\left(T,'Gauss', \gamma,C, \varepsilon_{1}, epoch_\alpha \right)$
\State \hspace{0.25cm}Compute $f^{'}(\gamma)$ by \eqref{formula_ex48}
\State \hspace{0.25cm}\textbf{If} $\left| f^{'}(\gamma) \middle| \leq \varepsilon_{2} \right.$
\State \hspace{0.5cm}\textbf{Break}
\State \hspace{0.25cm}\textbf{Else}
\State \hspace{0.5cm}Compute $\eta$ with the dynamic strategy
\State \hspace{0.5cm} $\gamma = \gamma + \eta f^{'}(\gamma)$
\State \hspace{0.25cm}\textbf{End If}
\State \textbf{End For}
\State $\gamma^{*} = \gamma$
\State \textbf{Return}
\end{algorithmic}
\label{alg3}
\end{algorithm}

\section{Experiments and discussions}
In order to explore the effectiveness of the MaxMin-L2-SVC-NCH model, this section compares the performance of MaxMin-L2-SVC-NCH with some previous best approaches on a series of public datasets. In addition, the training effectiveness of MaxMin-L2-SVC-NCH is tested on some representative datasets.

\subsection{The datasets}

\begin{table}[htbp]
  \centering
  \caption{The basic information of the datasets}
    \begin{tabular}{p{8.22em}p{5.665em}cc}
    \toprule
    Datasets & \multicolumn{1}{p{4.835em}}{Key}& \multicolumn{1}{p{4.835em}}{Instances}&\multicolumn{1}{p{4.78em}}{Attributes} \\
    \midrule
    Parkinsons & PAR   & 195   & 22 \\
    Sonar & SON   & 208   & 60 \\
    Spectf & SPE   & 267   & 44 \\
    Heart & HEA   & 270   & 13 \\
    Ionosphere & ION   & 351   & 33 \\
    Breast & BRE   & 569   & 30 \\
    Australian & AUS   & 690   & 14 \\
    German & GER   & 1000  & 24 \\
    Mushrooms & MUS   & 8124  & 112 \\
    Phishing & PHI   & 11055 & 68 \\
    \bottomrule
    \end{tabular}%
  \label{tab:tabel2}%
\end{table}%

The commonly used datasets are Parkinsons, Sonar, Spectf, Heart, Ionosphere, Breast, Australian, German, Mushrooms and Phishing, which are downloaded from LIBSVM datasets \cite{csie} and UCI datasets \cite{uci}. All datasets belong to two-classes problem. Table \ref{tab:tabel2} lists the basic information of the datasets.

\subsection{The experimental scheme} 
All models run in PyCharm 2021.2.3 on a DELL PC with i5-11400 2.60 GHz processors, 16.00 GB memory, and Windows 10.0 operating system. We compare some previous best approaches with our MaxMin-L2-SVC-NCH. The previous best approaches include the Bat-SVC, the commonly used CV-SVC, the Fisher-SVC and the LD-SVC. The Bat-SVC model, proposed by Yang et al., employs the bat algorithm to search for the optimal values of the Gaussian kernel parameter $\gamma$ and penalty parameter $C$\cite{yang2020nature}. The parameter setting of Bat-SVC refers to Yang's paper. The CV-SVC model uses the classic grid search approach to identify the best values for $\gamma$ and $C$. The Gaussian kernel parameter $\gamma$ and the penalty parameter $C$ are selected by 5-fold cross-validations in CV-SVC. The model is implemented by Sklearn Toolkit based on LIBSVM\cite{chang2011libsvm}. Fisher-SVC is a SVC model that employs fisher discriminant function to search for the optimal kernel parameters\cite{wang2003determination}. The parameter $C$ is obtained through grid search. Similarly, LD-SVC uses kernel density estimation to calculate in likelihood space, searching for the best parameters within a given range of Gaussian kernel parameters\cite{menezes2019width}. The penalty parameter $C$ is also obtained through grid search. MaxMin-L2-SVC-NCH is implemented by our GB algorithm. 

All datasets are standardized so that the mean value of each feature is 0 and the variance is 1 before the experiment. All datasets are randomly divided into training set and test set according to the ratio of 8:2. To eliminate the random effect of data division as much as possible, the random division is repeated thirty times. The mean and standard deviation (STD) of the classification accuracies over the thirty randomized trials are reported. Due to the large size of the last two datasets, MUS and PHI, only five repeated experiments were conducted for these two datasets.

In all previous best approaches, the candidate sets are $\gamma \in\{2^{-15}, 2^{-13}, \cdots, \\2^{3}\}$ and $C\in\{2^{-5}, 2^{-3}, \cdots, 2^{15}\}$, which are referenced from \cite{hsu2003practical}. In MaxMin-L2-SVC-NCH, $\gamma$ is selected by the gradient-based rule from $[2^{-15}, 2^{3}]$ that is the same as \cite{hsu2003practical},and the initial value is set by 0.004 that is the middle value between $2^{-15}$ and $2^{3}$. Choosing the middle value as the initial value is more conducive to find the best parameter. Unlike other baseline models, MaxMin-L2-SVC-NCH does not search for penalty parameters $C$. The parameter $C$ in MaxMin-L2-SVC-NCH is fixed as 1 \cite{menezes2019width} since it plays a small role. The other parameters $\varepsilon_{1}$, $\varepsilon_{2}$, $epoch_\gamma$, $epoch_\alpha$ in GB algorithm are set as $10^{-6}$, $10^{-3}$, 500 and 2000, respectively.

\subsection{The experimental results and discussions}

\subsubsection{Accuracy}
Table \ref{tab_3} presents the average accuracy and standard deviation on the test sets for the compared models and our MaxMin-L2-SVC-NCH. For the convenience of display, MaxMin-L2-SVC-NCH is shortened as MaxMin-SVC in tabels. Compared with other models, MaxMin-L2-SVC-NCH performs better overall and more stable on the ten datasets, while the other four baseline models have different performances. The MaxMin-L2-SVC obtains the best results 5 times on the ten datasets while the times of the best results for Bat-SVC, CV-SVC, Fisher-SVC and LD-SVC are 1, 1, 3, and 1, respectively.

To confirm above statement, we carry out a statistical hypothesis test with binomial cumulative distribution function. The random probability of the best among the five methods is 0.2. The P-value of MaxMin-L2-SVC in the statistical hypothesis test is 0.0064 that is less for 0.05. It indicates MaxMin-L2-SVC passes the statistical hypothesis test, i.e. the MaxMin-L2-SVC obtains the best performance among the five methods on the ten datasets.

\begin{table*}[htbp]\small
  \centering
  \caption{Accuracy results of the models}
    \begin{tabular}{p{3.5em}p{5.18em}p{5.18em}p{5.18em}p{5.18em}p{6.08em}}
    \toprule
    Datasets & Bat-SVC & CV-SVC & Fisher-SVC & LD-SVC & MaxMin-SVC \\
    \midrule
    PAR   & 76.15±1.64 & 90.43±4.77 & 86.84±3.42 & 88.21±4.00 & \textbf{90.68±3.89} \\
    SON   & 53.02±3.42 & 83.33±5.57 & 61.27±11.08 & 85.24±5.09 & \textbf{86.98±3.57} \\
    SPE   & \textbf{79.63±0.00} & 76.60±4.57 & \textbf{79.63±0.00} & 78.89±4.5 & 79.07±2.49 \\
    HEA   & 78.58±6.77 & \textbf{81.91±4.3} & 80.80±5.00 & 81.91±5.14 & 81.54±4.32 \\
    ION   & 90.38±7.34 & 92.21±2.79 & 93.15±5.65 & \textbf{94.37±2.57} & 91.88±2.29 \\
    BRE   & 63.16±0.0 & 96.61±1.46 & 96.02±1.98 & 96.64±1.79 & \textbf{97.40±1.31} \\
    AUS   & 85.24±3.23 & 85.19±2.64 & 83.28±3.33 & 85.34±2.92 & \textbf{85.39±2.87} \\
    GER   & 70.22±0.25 & 76.13±3.04 & \textbf{76.57±2.82} & 76.12±2.88 & 75.03±1.97 \\
    MUS   & 98.20±0.09 & 99.93±0.10 & \textbf{100.0±0.0} & 99.87±0.13 & 99.94±0.08 \\
    PHI   & 86.55±0.80 & 97.07±0.32 & 76.15±0.39 & 96.74±0.17 & \textbf{97.28±0.12} \\
    \bottomrule
    \end{tabular}%
  \label{tab_3}%
\end{table*}%

Bat-SVC is a genetic search algorithm that automatically searches for optimal parameters in the global space. Although Bat-SVC performed well on the SPE dataset with an accuracy rate of 79.63\%, it showed poor performance on other datasets and exhibited overall instability. The possible reason is that the solutions in each generation are hardly guaranteed to be excellent by Bat-SVC algorithm, which results in its unstable performance. On the other hand, CV-SVC performs well on most datasets and shows the best performance on the HEA dataset with an accuracy of 81.91\%. As a classic algorithm, CV-SVC has good experimental results but needs to train a large number of models. Its performance is usually relatively stable, and a large number of cross-validations can effectively reduce the generalization error. Fisher-SVC performs best on the SPE, GER, and MUS datasets, shows average performance on other datasets, and performs poorly on the SON and PHI datasets. Fisher-SVC uses the discriminator to optimize the within-class and inner-class distances, which requires extensive calculations. The optimization of distance to select parameters is also relatively unstable. Fisher-SVC is highly dependent on the distance of training samples distributed in different kernel spaces, which makes it easy to overfit the training data and is not conducive to generalization. Moreover, it is susceptible to noise data. LD-SVC performs best on the ION dataset and exhibits relatively stable performance overall. LD-SVC uses the kernel density estimation method to estimate the sample distribution in the likelihood space, making the calculation complexity higher since the calculation of the likelihood function is not easy. MaxMin-L2-SVC-NCH uses gradient information for training and shows excellent performance on most datasets, with relatively small standard deviations. Compared to other models, MaxMin-L2-SVC-NCH exhibits better overall accuracy and stability on the ten datasets. 

In summary, Bat-SVC utilizes the genetic algorithm to search for parameters globally, which makes it difficult to converge and results in unstable performance. On the other hand, the classic CV-SVC algorithm shows stable performance but requires a significant amount of time to search for optimal parameters. Fisher-SVC optimizes the distance of within-class and inner-class, which requires a large amount of calculation and exhibits unstable performance. The LD-SVC model uses the kernel density estimation method, which has high computational complexity. Finally, MaxMin-L2-SVC-NCH utilizes the gradient information and demonstrates excellent and stable performance.

\subsubsection{Training cost}
Regarding model training efficiency, CV-SVC is the only model that directly selects parameters through model training, while other models use different parameter search methods. Table \ref{tab_4} provides a comparison of the number of models trained. Grid search and 5-fold cross-validations in CV-SVC results in a total of 550 trained models, which is time-consuming. 

\begin{table}[htbp]
  \centering
  \caption{The number of trained models}
    \begin{tabular}{p{6.5em}c}
    \toprule
    Models & \multicolumn{1}{p{13.31em}}{The number of trained models} \\
    \midrule
    Bat-SVC    & 400 \\
    CV-SVC   & 550 \\
    Fisher-SVC   & 11 \\
    LD-SVC   & 11 \\
    MaxMin-SVC & \textbf{8.2} \\
    \bottomrule
    \end{tabular}%
  \label{tab_4}%
\end{table}%

The Bat-SVC algorithm employs 20 bats to select parameters through a genetic algorithm and iterates 20 times, thus training a total of 400 SVM models. Fisher-SVC pre-searches the Gaussian kernel parameters through the discriminator, where the intra-class distance and inter-class distance of the sample is needed to calculate and the training of the model is not involved here. However, the discriminator involves significant computation cost and memory usage. The penalty parameter $C$ is selected by grid search, so 11 models need to be trained in Fisher-SVC. The method of selecting parameters in LD-SVC is similar to the Fisher-SVC algorithm, the difference is that the LD-SVC algorithm replaces the discriminator with a kernel density estimator. The parameter $C$ in LD-SVC is also selected by grid search, so 11 models need to be trained. In addition, the kernel density estimation also requires additional computation cost and memory usage.      

The MaxMin-L2-SVC-NCH model selects Gaussian kernel parameters directly via gradient information. The number of trained models depends on the number of iterations. The average number of iterations is only 8.2 in MaxMin-L2-SVC-NCH and the number of trained models is the least. MaxMIn-L2-SVC-NCH does not require heuristic search of Gaussian kernel parameters in advance, and it only needs to calculate the gradient information of the Gaussian kernel during model training and use the dynamic learning rate to quickly update the Gaussian kernel parameters. In addition, MaxMin-L2-SVC-NCH does not require the grid searches for the parameter $C$. In MaxMin-L2-SVC-NCH, we evaluate different Gaussian kernel parameters by a fixed $C$ from \eqref{formula_ex7}. It is fair because the weights of two items in the objective function for different Gaussian kernel parameters is the same.  However,  CV-SVC, Bat-SVC, Fisher-SVC and LD-SVC can not do that as MaxMin-L2-SVC-NCH.  \\

\subsubsection{The effectiveness of training MaxMin-L2-SVC-NCH}
To analyze the training process of MaxMin-L2-SVC-NCH, we provide the changes of $f'(\gamma)$, the ratio of the inter-class distance to the intra-class distance on four representative datasets: Parkinsons, Heart, Germen and Mushrooms. Parkinsons, Heart and Mushrooms are natural data while Germen are economic data. The sizes of Parkinsons, Heart and Germen are small while the size of Mushrooms is relatively large.
\begin{figure}[H]
\centering
\includegraphics{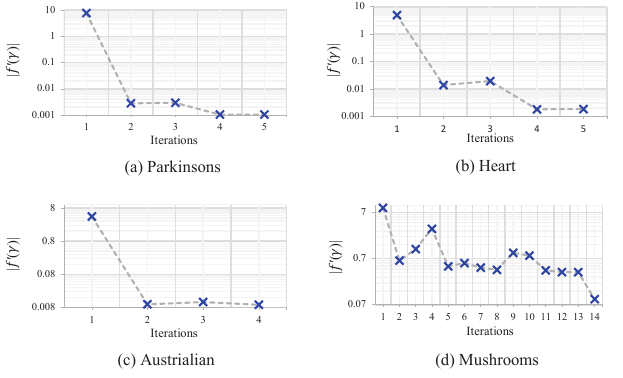}
\caption{The changes of $f'(\gamma)$ on the representative datasets. The vertical coordinate is logarithmic coordinate.}
\label{fig_1}
\end{figure}

Fig.\ref{fig_1} shows the changes of $f'(\gamma)$ during the training of MaxMin-L2-SVC-NCH on the representative datasets. From Fig.\ref{fig_1}, we see that all $f'(\gamma)$ quickly converge to zero with the required accuracy after a small number of iterations. It indicates that the training process of MaxMin-L2-SVC-NCH is efficient.

\begin{figure}[H]
\centering
\includegraphics{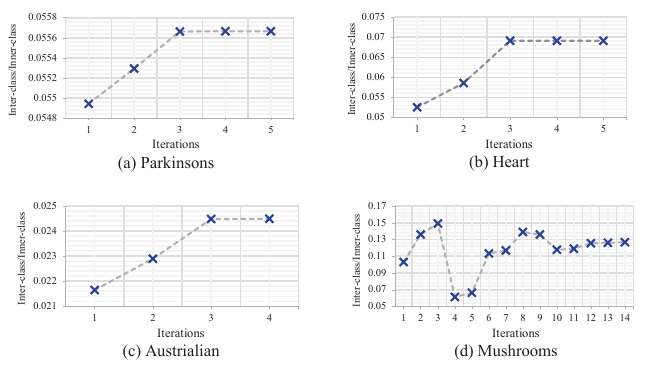}
\caption{The ratios of inter-class distance and intra-class distance on the representative datasets.}
\label{fig_2}
\end{figure}

Fig.\ref{fig_2} provides the ratio of the inter-class distance to the intra-class variance, i.e. Fisher discriminant function \cite{wang2003determination}. The inter-class distance is defined by
\begin{multline}
\text{$D_{Inter} = \left\| {m^{+}{- m}^{-}} \right\|^{2}$}\\
\text{$=\left( \frac{1}{l^{+}} \right)^{2}{\sum\limits_{i = 1}^{l^{+}}{\sum\limits_{j = 1}^{l^{+}}{k\left( {\boldsymbol{x}_{i},\boldsymbol{x}_{j}} \right)}}} + \left( \frac{1}{l^{-}} \right)^{2}{\sum\limits_{i = 1}^{l^{-}}{\sum\limits_{j = 1}^{l^{-}}{k\left( {\boldsymbol{x}_{i},\boldsymbol{x}_{j}} \right)}}}$}
\text{$- \frac{2}{l^{+}l^{-}}{\sum\limits_{i = 1}^{l^{+}}{\sum\limits_{j = 1}^{l^{-}}{k\left( {\boldsymbol{x}_{i},\boldsymbol{x}_{j}} \right)}}}$}
\end{multline}where $m^{+} = \frac{1}{l^{+}}{\sum\limits_{i = 1}^{l^{+}}{\Phi\left( \boldsymbol{x}_{i} \right)}}$ and $m^{-} = \frac{1}{l^{-}}{\sum\limits_{i = 1}^{l^-}{\Phi\left( \boldsymbol{x}_{i} \right)}}$. The intra-class variance is defined by
\begin{equation}
\label{formula_ex51}
D_{Inner} = \frac{1}{l^{+}}{\sum\limits_{i = 1}^{l^{+}}\left\| {m^{+} - \Phi\left( \boldsymbol{x}_{i} \right)} \right\|^{2}} + \frac{1}{l^{-}}{\sum\limits_{i = 1}^{l^{-}}\left\| {m^{-} - \Phi\left( \boldsymbol{x}_{i} \right)} \right\|^{2}}.
\end{equation}
From Fig.\ref{fig_2}, we see that the ratios of the inter-class distance to the intra-class variance all converge to relatively large values. It indicates that our algorithm also obtains a similar effect of maximizing the ratio of the inter-class distance to the intra-class variance.

\subsubsection{A evaluation to the selected parameters}
The above experiments have shown that the effectiveness and the superiority of MaxMin-L2-SVC-NCH in selecting Gaussian kernel parameter $\gamma$. What happens to the compared methods if they simply use the parameter selected by MaxMin-L2-SVC-NCH? In reverse, what happens to MaxMin-L2-SVC-NCH if it simply uses the parameters from the compared methods or take them as the initial values. These problems are interesting. Each model has its own method of finding parameters. In general, the selected parameters are evaluated by fair experiment results. It is rare to apply the parameters selected by a certain model directly to the comparison models. However, it provides another perspective on evaluating the selected parameters. Table \ref{tab_5} lists the results on the dataset PAR if MaxMin-L2-SVC-NCH simply uses the parameters selected by the compared methods while Table \ref{tab_6} lists the reverse results. In Table \ref{tab_5} and \ref{tab_6}, $\text{Test\_acc}$ denotes the test accuracy where $\text{Test\_acc1}$ and $\text{Test\_acc2}$ respectively indicate the test accuracies of MaxMin-L2-SVC-NCH when the $\gamma$ selected by the compared methods is simply used as the given value or the initial value. In Table \ref{tab_6}, the accuracies of Bat-SVC, CV-SVC, Fisher-SVC and LD-SVC are the same because they all use the standard SVC as the classification model. The $C$ in all methods is set to 1 for fair comparison.

From Table \ref{tab_5}, we see that the accuracies of MaxMin-L2-SVC-NCH are not inferior to the compared methods if it simply uses the $\gamma$ selected by the compared methods. Table \ref{tab_6} indicates that the $\gamma$ selected by MaxMin-L2-SVC-NCH is also good for the standard SVC. These results further show that the parameter selected by MaxMin-L2-SVC-NCH is not only good for itself but also good for the compared methods.
% \usepackage{booktabs}
\begin{table}[]
\centering
\caption{The accuracies of MaxMin-L2-SVC-NCH if it simply uses the parameters selected by the compared methods}
\begin{tabular}{@{}ccccc@{}}
\toprule
\multicolumn{3}{c}{The compared methods}    & \multicolumn{2}{c}{MaxMin-SVC}  \\ 
\cmidrule(lr){4-5} \cmidrule(lr){1-3}
\multicolumn{1}{l}{Models}
           & The selected $\gamma$ & Test\_acc(\%) & Test\_acc1(\%) & Test\_acc2(\%) \\ \midrule
Bat-SVC    & 8              & 74.36         & 74.36          & 74.36          \\
CV-SVC     & 0.125          & 92.31         & 92.31          & 92.31          \\
Fisher-SVC & 3.05E-05       & 74.36         & 74.36          & 74.36          \\
LD-SVC     & 0.03125        & 87.18         & 92.31          & 94.87          \\ \bottomrule
\label{tab_5}
\end{tabular}
\end{table}
% \usepackage{booktabs}
% \usepackage{multirow}
\begin{table}[]
\centering
\caption{The accuracies of the compared methods if they simply use the parameter selected by MaxMin-L2-SVC-NCH}
\begin{tabular}{@{}cccc@{}}
\toprule
\multicolumn{2}{c}{The compared methods} & \multicolumn{2}{c}{MaxMin-SVC}                    \\ \cmidrule(lr){1-2} \cmidrule(lr){3-4}
\multicolumn{1}{l}{Models} & \multicolumn{1}{l}{Test\_acc(\%)} & \multicolumn{1}{l}{The selected $\gamma$} & \multicolumn{1}{l}{Test\_acc(\%)} \\ \midrule
Bat-SVC       & \multirow{4}{*}{92.31}   & \multirow{4}{*}{0.24995} & \multirow{4}{*}{92.31} \\
CV-SVC        &                          &                          &                        \\
Fisher-SVC    &                          &                          &                        \\
LD-SVC        &                          &                          &                        \\ \bottomrule
\label{tab_6}
\end{tabular}
\end{table}

\section{Conclusions}
A novel method, MaxMin-L2-SVC-NCH, is proposed for training SVC and selecting Gaussian kernel parameters. MaxMin-L2-SVC-NCH is a minimax problem, where the minimization problem is L2-SVC-NCH and the maximization problem aims to select the optimal Gaussian kernel parameters. A lower time complexity is expected in MaxMin-L2-SVC-NCH because the time-consuming CV is not needed. To solve L2-SVC-NCH quickly and efficiently, the PGA is proposed. The PGA provides more flexibility than the famous SMO algorithm since the SMO can be viewed as a special case of the PGA. For the solution of the maximization problem, the GA-DLR algorithm is proposed. A gradient-based algorithm is subsequently provided for the solution of MaxMin-L2-SVC-NCH by connecting the PGA and the GA-DLR algorithm in series. Experimental results on the public datasets reveal that MaxMin-L2-SVC-NCH significantly reduces the number of trained models while maintaining competitive testing accuracy compared to the previous best approaches. These findings reveal an enhanced performance of MaxMin-L2-SVC-NCH, which indicates it may be a better choice for SVC tasks.

It is a future work to optimize the implementation of MaxMin-L2-SVC-NCH so that it is suitable for very large datasets. Additionally, investigating the impact of the initial value of the kernel parameter $\gamma$ on MaxMin-L2-SVCNCH is worthy of further exploration. Furthermore, extending MaxMin-L2-SVC-NCH to encompass multi-class and regression problems is also an interesting task.

\section*{Data availability}
 All data used are public datasets that are available from the references. The source code for MaxMin-L2-SVC-NCH can be found at \href{https://github.com/visitauto/MaxMin-L2-SVC-NCH}{https://github\\.com/visitauto/MaxMin-L2-SVC-NCH}.
\section*{Acknowledgement}
This work is supported in part by the National Natural Science Foundation of China under grant No.62171391. We also thank Shaorong Fang and Tianfu Wu from Information and Network Center of Xiamen University for the help with the GPU computing.

\bibliographystyle{elsarticle-num} 
\bibliography{references}

\begin{thebibliography}{10}
\expandafter\ifx\csname url\endcsname\relax
  \def\url#1{\texttt{#1}}\fi
\expandafter\ifx\csname urlprefix\endcsname\relax\def\urlprefix{URL }\fi
\expandafter\ifx\csname href\endcsname\relax
  \def\href#1#2{#2} \def\path#1{#1}\fi

\bibitem{1995Support}
C.~Cortes, V.~Vapnik, Support-vector networks, Machine learning 20~(3) (1995) 273--297.

\bibitem{vidic2018support}
I.~Vidi{\'c}, L.~Egnell, N.~P. Jerome, J.~R. Teruel, T.~E. Sj{\o}bakk, A.~{\O}stlie, H.~E. Fj{\o}sne, T.~F. Bathen, P.~E. Goa, Support vector machine for breast cancer classification using diffusion-weighted mri histogram features: Preliminary study, Journal of Magnetic Resonance Imaging 47~(5) (2018) 1205--1216.

\bibitem{sethy2020deep}
P.~K. Sethy, N.~K. Barpanda, A.~K. Rath, S.~K. Behera, Deep feature based rice leaf disease identification using support vector machine, Computers and Electronics in Agriculture 175 (2020) 105527.

\bibitem{sun2017dynamic}
J.~Sun, H.~Fujita, P.~Chen, H.~Li, Dynamic financial distress prediction with concept drift based on time weighting combined with adaboost support vector machine ensemble, Knowledge-Based Systems 120 (2017) 4--14.

\bibitem{PAN201790}
Y.~Pan, Z.~Xiao, X.~Wang, D.~Yang, A multiple support vector machine approach to stock index forecasting with mixed frequency sampling, Knowledge-Based Systems 122 (2017) 90--102.

\bibitem{hoang2018image}
N.-D. Hoang, Q.-L. Nguyen, D.~Tien~Bui, Image processing--based classification of asphalt pavement cracks using support vector machine optimized by artificial bee colony, Journal of Computing in Civil Engineering 32~(5) (2018) 04018037.

\bibitem{fan2019research}
Z.~Fan, C.~Liu, D.~Cai, S.~Yue, Research on black spot identification of safety in urban traffic accidents based on machine learning method, Safety science 118 (2019) 607--616.

\bibitem{manavalan2017svmqa}
B.~Manavalan, J.~Lee, Svmqa: support--vector-machine-based protein single-model quality assessment, Bioinformatics 33~(16) (2017) 2496--2503.

\bibitem{ruan2021convex}
Y.~Ruan, Y.~Xiao, Z.~Hao, B.~Liu, A convex model for support vector distance metric learning, IEEE Transactions on Neural Networks and Learning Systems 33~(8) (2021) 3533--3546.

\bibitem{avolio2020semiproximal}
M.~Avolio, A.~Fuduli, A semiproximal support vector machine approach for binary multiple instance learning, IEEE transactions on neural networks and learning systems 32~(8) (2020) 3566--3577.

\bibitem{tang2021valley}
L.~Tang, Y.~Tian, W.~Li, P.~M. Pardalos, Valley-loss regular simplex support vector machine for robust multiclass classification, Knowledge-Based Systems 216 (2021) 106801.

\bibitem{li2021dc}
G.~Li, L.~Yang, Z.~Wu, C.~Wu, Dc programming for sparse proximal support vector machines, Information Sciences 547 (2021) 187--201.

\bibitem{zhang2019optimal}
T.~Zhang, Z.-H. Zhou, Optimal margin distribution machine, IEEE Transactions on Knowledge and Data Engineering 32~(6) (2019) 1143--1156.

\bibitem{wang2003determination}
W.~Wang, Z.~Xu, W.~Lu, X.~Zhang, Determination of the spread parameter in the gaussian kernel for classification and regression, Neurocomputing 55~(3-4) (2003) 643--663.

\bibitem{tsang2005core}
I.~W. Tsang, J.~T. Kwok, P.-M. Cheung, N.~Cristianini, Core vector machines: Fast svm training on very large data sets, Journal of Machine Learning Research 6~(4) (2005).

\bibitem{sun2010analysis}
J.~Sun, C.~Zheng, X.~Li, Y.~Zhou, Analysis of the distance between two classes for tuning svm hyperparameters, IEEE transactions on neural networks 21~(2) (2010) 305--318.

\bibitem{menezes2019width}
M.~V. Menezes, L.~C. Torres, A.~P. Braga, Width optimization of rbf kernels for binary classification of support vector machines: A density estimation-based approach, Pattern Recognition Letters 128 (2019) 1--7.

\bibitem{akram2022fast}
Z.~Akram-Ali-Hammouri, M.~Fern{\'a}ndez-Delgado, E.~Cernadas, S.~Barro, Fast support vector classification for large-scale problems, IEEE Transactions on Pattern Analysis and Machine Intelligence 44~(10) (2022) 6184--6195.

\bibitem{FRIEDRICHS2005107}
F.~Friedrichs, C.~Igel, Evolutionary tuning of multiple svm parameters, Neurocomputing 64 (2005) 107--117.

\bibitem{tharwat2017ba}
A.~Tharwat, A.~E. Hassanien, B.~E. Elnaghi, A ba-based algorithm for parameter optimization of support vector machine, Pattern recognition letters 93 (2017) 13--22.

\bibitem{peng2011soft}
H.~Peng, M.~Luo, W.~Lv, L.~Luo, A soft-margin support vector machine based on normal convex hulls, International Journal of Advancements in Computing Technology 3~(7) (2011).

\bibitem{bennett2000duality}
K.~P. Bennett, E.~J. Bredensteiner, Duality and geometry in svm classifiers, in: Proceedings of 17th International Conference on Machine Learning, 2000, pp. 57--64.

\bibitem{1998Sequential}
J.~Platt, Sequential minimal optimization: A fast algorithm for training support vector machines, Advances in Kernel Methods-Support Vector Learning 208 (07 1998).

\bibitem{keerthi2001improvements}
S.~S. Keerthi, S.~K. Shevade, C.~Bhattacharyya, K.~R.~K. Murthy, Improvements to platt's smo algorithm for svm classifier design, Neural computation 13~(3) (2001) 637--649.

\bibitem{fan2005working}
R.-E. Fan, P.-H. Chen, C.-J. Lin, T.~Joachims, Working set selection using second order information for training support vector machines., Journal of machine learning research 6~(12) (2005) 1889--1918.

\bibitem{2005optimization}
B.~Chen, Optimization Theory and Algorithms (Second Edition), Tsinghua University Press, 2005.

\bibitem{Linear}
D.~Luenberger, Y.~Ye, Linear and Nonlinear Programming (Third Edition), Springer, 2008.

\bibitem{csie}
C.-C. Chang, Libsvm data: Classification, regression, and multi-label, \url{https://www.csie.ntu.edu.tw/~cjlin/libsvmtools/datasets/} (2008).

\bibitem{uci}
K.~Markelle, L.~Rachel, N.~Kolby, The uci machine learning repository, \url{https://archive.ics.uci.edu/} (2007).

\bibitem{yang2020nature}
X.-S. Yang, Nature-inspired optimization algorithms, Academic Press, 2020.

\bibitem{chang2011libsvm}
C.-C. Chang, C.-J. Lin, Libsvm: A library for support vector machines, ACM Transactions on Intelligent Systems and Technology 2~(3) (2011) 1--27.

\bibitem{hsu2003practical}
C.-W. Hsu, C.-C. Chang, C.-J. Lin, et~al., A practical guide to support vector classification (2003).

\end{thebibliography}
% \end{thebibliography}

\end{document}